\documentclass[10pt,twocolumn,letterpaper]{article}

\usepackage{cvpr}              % To produce the CAMERA-READY version
\usepackage{microtype}

\definecolor{cvprblue}{rgb}{0.21,0.49,0.74}
\usepackage[pagebackref,breaklinks,colorlinks,allcolors=cvprblue]{hyperref}

\def\paperID{20759} % *** Enter the Paper ID here
\def\confName{CVPR}
\def\confYear{2026}

\title{IMAIA: Interactive Maps AI Assistant for Travel Planning and Geo-Spatial Intelligence}

\author{Jieren Deng,
Zhizhang Hu\thanks{Zhizhang contributed to this work while he was with Microsoft},
Ziyan He,
Aleksandar Cvetkovic,\\
Pak Kiu Chung,
Dragomir Yankov,
Chiqun Zhang\\
Microsoft, Amazon\\
{\tt\scriptsize \{jierendeng, ziyanhe, acvetkovic\}@microsoft.com}\\
{\tt\scriptsize \{pachung, dragoy, chizhang\}@microsoft.com}\\
{\tt\scriptsize zhizhanh@amazon.com}
}
\begin{document}
\maketitle
% \begin{abstract}
% Map applications are still largely point-and-click, making it difficult to ask map-centric questions or connect what a camera sees to the surrounding geospatial context with view-conditioned. We introduce IMAIA, an interactive Maps AI Assistant that enables natural-language interaction with both vector (street) maps and satellite imagery, and augments camera inputs with geospatial intelligence to help users understand the world. IMAIA comprises two complementary components. Maps Plus treats the map as first-class context by parsing tiled vector/satellite views into a grid-aligned representation that a language model can query to resolve deictic references (e.g., “the flower-shaped building next to the park in the top-right”). Places AI Smart Assistant (PAISA) performs camera-aware place understanding by fusing image–place embeddings with geospatial signals (location, heading, proximity) to ground a scene, surface salient attributes, and generate concise explanations. A lightweight multi-agent design keeps latency low and exposes interpretable intermediate decisions. Across map-centric QA and camera-to-place grounding tasks, IMAIA improves accuracy and responsiveness over strong baselines while remaining practical for user-facing deployments. By unifying language, maps, and geospatial cues, IMAIA moves beyond scripted tools toward conversational mapping that is both spatially grounded and broadly usable.

% \end{abstract}

\begin{abstract}
Modern mapping tools remain fundamentally point–and–click, offering little support for interactive, multimodal engagement with either map views or real-world camera inputs. This gap becomes most apparent in the last-100-meter of navigation and during in-situ local discovery, where users naturally combine spatial reasoning over maps with ego-centric perception of their surroundings.
We introduce \textbf{IMAIA}, an interactive geospatial assistance framework that unifies three core capabilities: (1) map-centric spatial understanding, (2) camera-to-place grounding, and (3) human-centered, bearing-aware navigation. IMAIA is composed of two interoperable components—\textbf{Maps Plus} and \textbf{PAISA}—coordinated by a lightweight multi-agent orchestrator. Maps Plus converts vector and satellite maps into a grid-aligned representation that enables flexible, view-conditioned spatial queries. PAISA fuses camera imagery with geospatial signals such as location, heading, and proximity to interpret ego-centric scenes and surface relevant attributes of nearby places.
The framework is fully modular: vision-language backends are replaceable without changing system behavior. IMAIA provides responsive, interpretable, and context-aware assistance across both map-based and real-world interactions, enabling a new class of interactive mapping experiences that remain robust and practical for user-facing deployment.
\end{abstract}
    
\section{Introduction}
\label{sec:intro}
\begin{figure*}[t]
    \centering
    \includegraphics[width=0.80\textwidth]{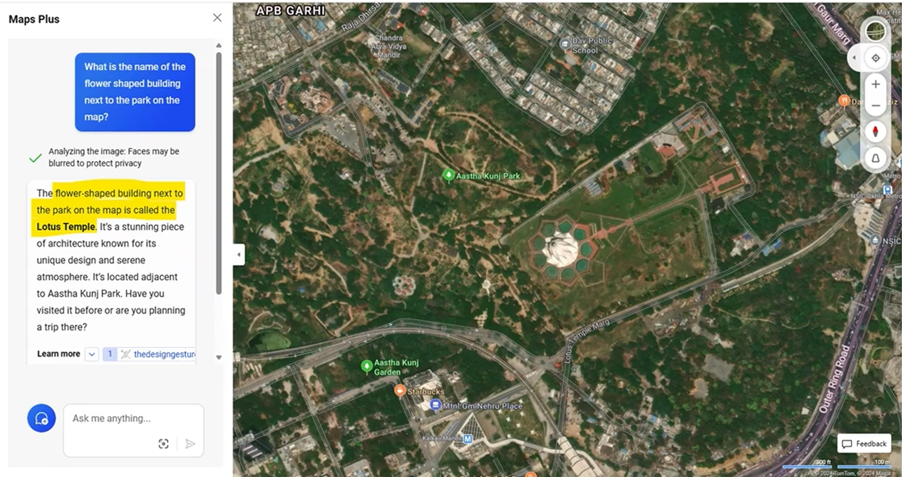}  % Adjust width as needed
    \caption{\textbf{User interface of Maps Plus} showing handling a query  \textit{``What is the name of the flower-shaped building next to the park on the map''} from the user.}
    \label{fig:maps_plus_ui}
\end{figure*}
Modern map applications remain largely point–and–click: users pan and zoom, then issue basic, limited, and inflexible queries. This interaction model breaks down for \emph{map-centric, view-conditioned} queries—e.g., “What is the flower-shaped building next to the park in the top-right of what I am viewing?”—and for connecting what a mobile camera sees to the surrounding geospatial context. The desire to explore and navigate unfamiliar environments is fundamental, yet current tools remain inadequate: traditional travel and mapping systems, constrained by static methodologies, struggle with real-world dynamism—fluctuating conditions, imprecise signals, and unexpected disruptions that degrade the experience. Meanwhile, travel planning, navigation, and local discovery are typically engineered as isolated modules, yielding fragmented interactions and brittle handoffs \cite{zhang2020building, dickinson2014tourism}; prior work even documents user behavior under disrupted plans \cite{li2020exploring}, yet practical systems still fail to support coherent, interactive map understanding at the moment it is most needed.

Recent advances in Large Language Models (LLMs) offer a path to transcend these limitations \cite{liu2024toward, zhang2024context, chen2024travelagent}. LLMs can process and synthesize diverse multimodal inputs, including text, imagery, geospatial data, and contextual cues \cite{rocchietti2024geolocated, hu2023provla,hu2024noised}, enabling more cohesive and adaptive geospatial assistants \cite{zhang2023map}. Two trends are particularly relevant. First, LLMs are increasingly adept at interpreting unstructured or ambiguous geospatial information, transforming vague user requests into precise map coordinates or actionable insights from noisy datasets \cite{zhang2023map}. Second, advancements in interactive and conversational AI highlight the need for multi-turn, context-aware interactions in navigation and discovery, where goals evolve over time \cite{yi2024survey}. In parallel, vision–language models (VLMs~\cite{10445007}) can describe images, but without explicit grounding to the current map state (viewport, scale, nearby entities) and geospatial signals (location, heading, proximity), their reasoning remains fragile in the presence of ambiguous visual cues or noisy inputs.~\cite{11228719, deng-etal-2021-tag-gradient}. Prior work has shown that transferring knowledge from larger models to smaller task-specialized models through distillation can improve robustness and efficiency \cite{DENG202512,Deng2023SmoothAS}, motivating our use of a distilled spatial reasoning module in IMAIA.

Against this backdrop, we focus on a concrete application problem: \emph{interactive, multimodal interaction with maps and real-world ego-centric views that bridges desktop trip planning and in-situ, last-100-meter navigation and local discovery}. We target users in two complementary settings: (i) planning and exploring on a desktop or phone map, where they want to reason about what they see on vector and satellite views, and (ii) arriving near their destination, where they want to understand what their camera is facing in the real world and receive guidance that respects their current bearing and surroundings. To support this end-to-end experience, our system must unify three capabilities that are typically handled in isolation:
(1) \textbf{map-centric spatial understanding}, where the map viewport is encoded as a quadkey-indexed grid and exposed as a structured visual prompt for interactive, view-conditioned reasoning over vector and satellite maps;
(2) \textbf{ego-centric scene understanding and geospatial place grounding}, where first-person camera views are connected to nearby entities using both visual features and geospatial context—implemented through our PAISA module; and
(3) \textbf{human-centered navigation}, providing bearing-aware guidance that matches how people naturally orient themselves and move through space in the last-100-meter phase.

In this paper, we introduce the \textbf{Interactive Maps AI Assistant (IMAIA)}, a modular framework built around two tightly integrated components---\textbf{Maps Plus} and the \textbf{Places AI Smart Assistant (PAISA)}---coordinated by a lightweight multi-agent orchestration layer. At the system and application level, Maps Plus delivers an interactive map experience for desktop-style exploration over vector and satellite maps, while PAISA provides an AR-style experience for interacting with real-world scenes grounded in associated geospatial data. Concretely, IMAIA operationalizes the three capabilities above as follows:

\begin{itemize}

\item \textbf{Maps Plus: interactive map experience via quadkey-based visual prompting.}
We introduce Maps Plus, an interactive map interface that enables view-conditioned, map-centric queries directly over vector and satellite maps. The system converts the current viewport into a quadkey-indexed grid and attaches visual and semantic attributes to each tile, forming a structured visual prompt that supports deictic spatial reasoning during map exploration.

\item \textbf{PAISA: ego-centric scene understanding and geospatial place grounding.}
We introduce PAISA, a camera-based module that grounds first-person views to nearby places by fusing visual input with geospatial signals such as location, heading, and proximity. Within a multi-agent architecture for orchestration, location intelligence, navigation, and spatial reasoning, PAISA connects observed scenes to surrounding entities and generates human-centered, bearing-aware navigation guidance.

\item \textbf{IMAIA: a unified multimodal framework with strong empirical performance.}
To the best of our knowledge, IMAIA is the first framework that tightly couples interactive map exploration with camera-grounded place understanding and last-100-meter navigation within a unified multimodal AI assistant. The framework is model-agnostic and modular, enabling interchangeable VLM and vision backends. Empirically, Maps Plus improves place detection accuracy from under 43\% to nearly 90\%, while our spatial reasoning model achieves 84\% accuracy with a 7.3$\times$ inference speedup over agent-based pipelines.

\end{itemize}
\section{Related Work}
\label{sec:maisa}
\label{sec:related-work}
\subsection{Geospatial/Maps Intelligence with LLMs}
Recent advancements in Large Language Models (LLMs) \cite{touvron2023llama,mao2023large} are driving a paradigm shift in information retrieval, moving from single-query, text-only systems to conversational, multi-modal search. This trend is particularly prominent in the geospatial domain, where models have been empowered with grounded language understanding \cite{li2023geolm}. A growing body of work has demonstrated the potential of LLMs for enriched map searches, adeptly handling spatio-temporal data and conversational queries \cite{zhang2024context, zhang2023map, Wazzan_2024, liu2024large, roberts2023gpt4geo}. However, a key challenge remains largely unaddressed: how to efficiently feed information from existing geo-indexing systems into an LLM. Little research has focused on this critical interface, which is necessary for the performant resolution of complex, multi-modal geospatial queries.
\subsection{Spatial Intelligence and Reasoning with VLMs}
Recent works have attempted to augment vision–language models (VLMs) with spatial reasoning capabilities. ASMv2~\cite{wang2024all} introduces fine-tuned modules for spatial VQA, SpatialVLM~\cite{chen2401spatialvlm} synthesizes large-scale spatial question–answer pairs to improve metric distance estimation, and SpatialRGPT~\cite{cheng2024spatialrgpt} incorporates scene graphs for relational reasoning. While these approaches advance spatial intelligence, they often exhibit two key limitations: (i) high latency, due to reliance on computationally intensive synthetic pipelines or graph-based reasoning, and (ii) task misalignment, as training objectives are primarily benchmark-driven rather than optimized for embodied agent tasks.
\section{Interactive Maps AI Assistant}
\subsection{Maps Plus}
\begin{figure}[htbp!]
    \centering
    \includegraphics[width=\linewidth]{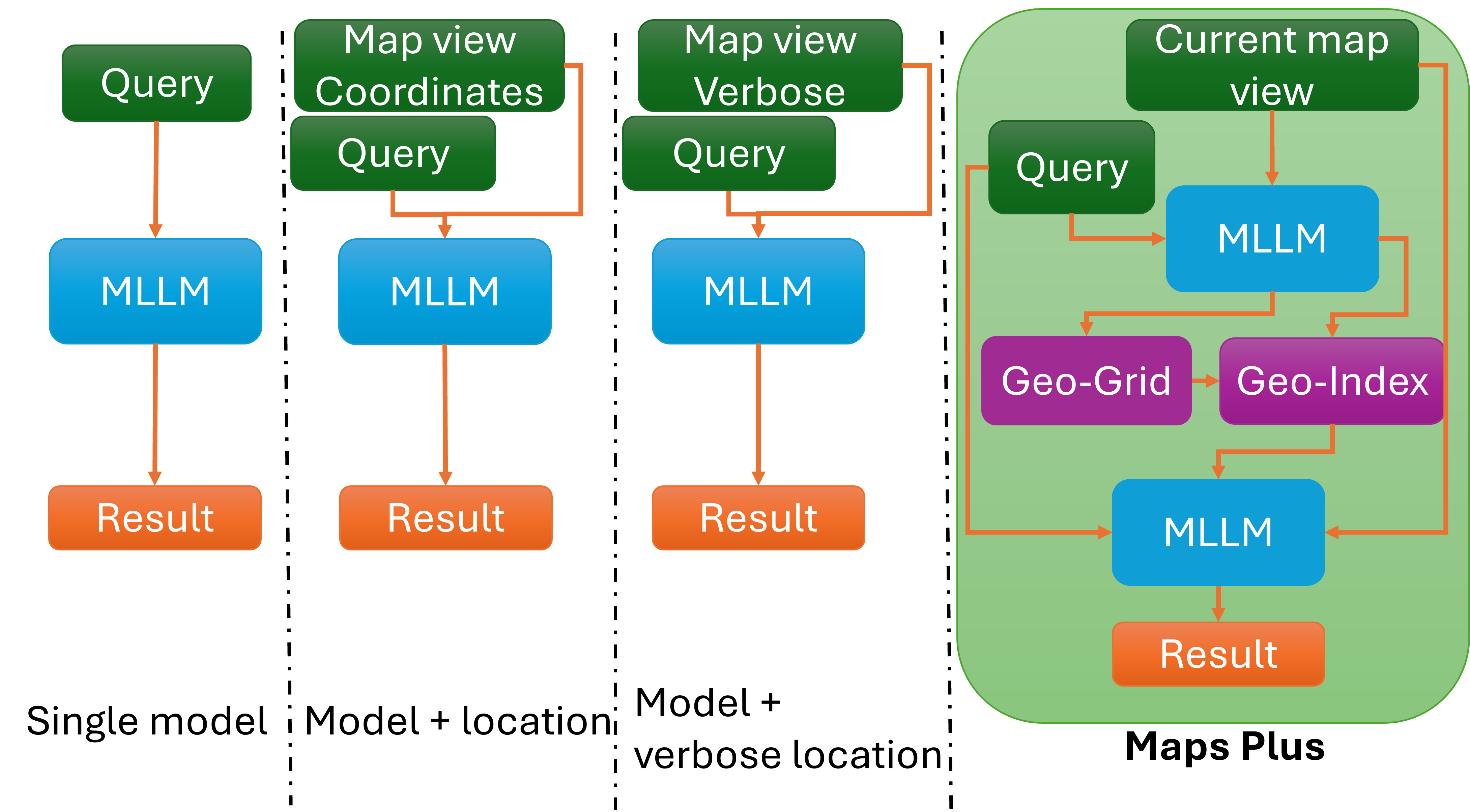}  % Adjust width as needed
    \caption{\textbf{Workflow comparison} of four settings: a standalone MLLM, an MLLM with coordinates, an MLLM with verbose place context, and our Maps Plus approach.}
    \label{fig:maps_plus_comp}
\end{figure}
Travelers usually begin by asking broad questions such as “What are the top locations to visit in X?”, “What should I see in X?”, or “Plan me a route through X.” 
Previous studies have shown that a large language model (LLM) can answer such queries either directly from its internal memory or by coordinating with external tools~\cite{zhang2023map, zhang2024context}, as illustrated in Figure~\ref{fig:maps_plus_comp}. 
These approaches are referred to as \textit{Single Model}, \textit{Model + Location}, and \textit{Model + Verbose Location}.
Once users begin examining an interactive map, however, they often pose richer, map-centric questions that require spatial reasoning about what they are actually viewing rather than a static list of attractions.
To meet these needs, we introduce a multimodal system that blends an LLM with image input and geospatial search (as shown in Figure~\ref{fig:maps_plus_ui}).
A user can click on map tiles or satellite imagery and converse naturally about that view, receiving answers grounded in both textual knowledge and visual context. 
As shown in Figure~\ref{fig:maps_plus_comp}, our proposed method first determines the geographic focus and zoom level of the user’s current view, then scans the surrounding imagery on a regular grid to extract visual features and detect salient geographic entities. 
Finally, it queries a geospatial index with those entities and synthesizes the results so the LLM can craft an informed, location-aware response. 
By treating the map itself as conversational context, the system supports fluid trip planning and exploratory tasks that go well beyond what text-only approaches can deliver.
\subsubsection{Location Awareness}
The first step involves providing GPT-4o with contextual information about where an image was taken. This can be presented in a structured format, such as precise latitude and longitude coordinates (e.g., 42.344, 36.236) or as a verbose description of the place (e.g., Seattle, WA, USA). This location-aware capability allows the model to ground its responses in geographic context, ensuring more relevant and accurate interpretations of map-based queries.
\begin{figure}[htbp!]
    \centering
    \includegraphics[width=1\linewidth]{
    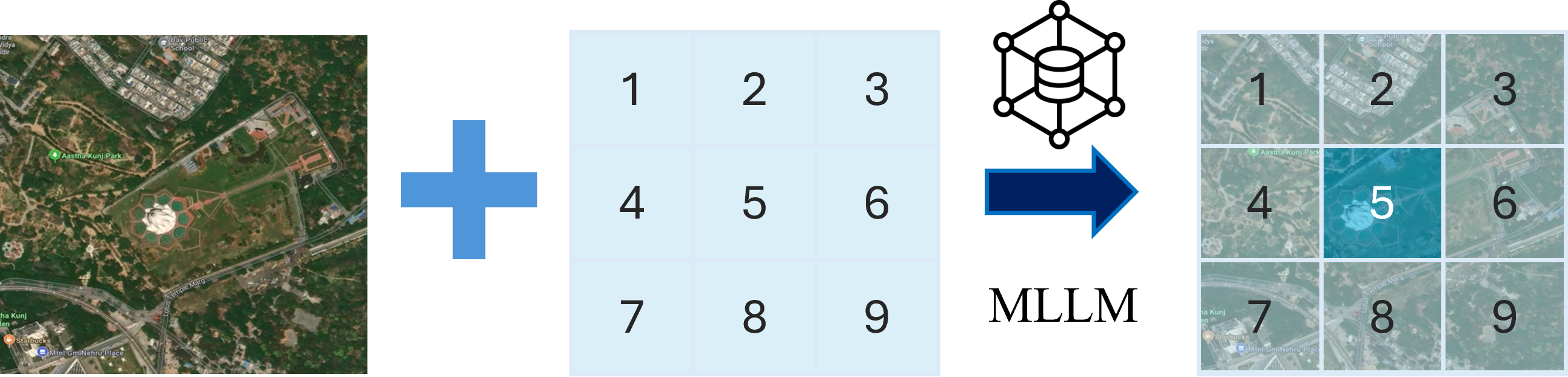}
    \caption{Illustration of quadkey-based visual prompting.}
    \label{fig:grid-layout}
\end{figure}

\begin{figure*}[htbp!]
    \centering
    \includegraphics[width=0.9\textwidth]{
    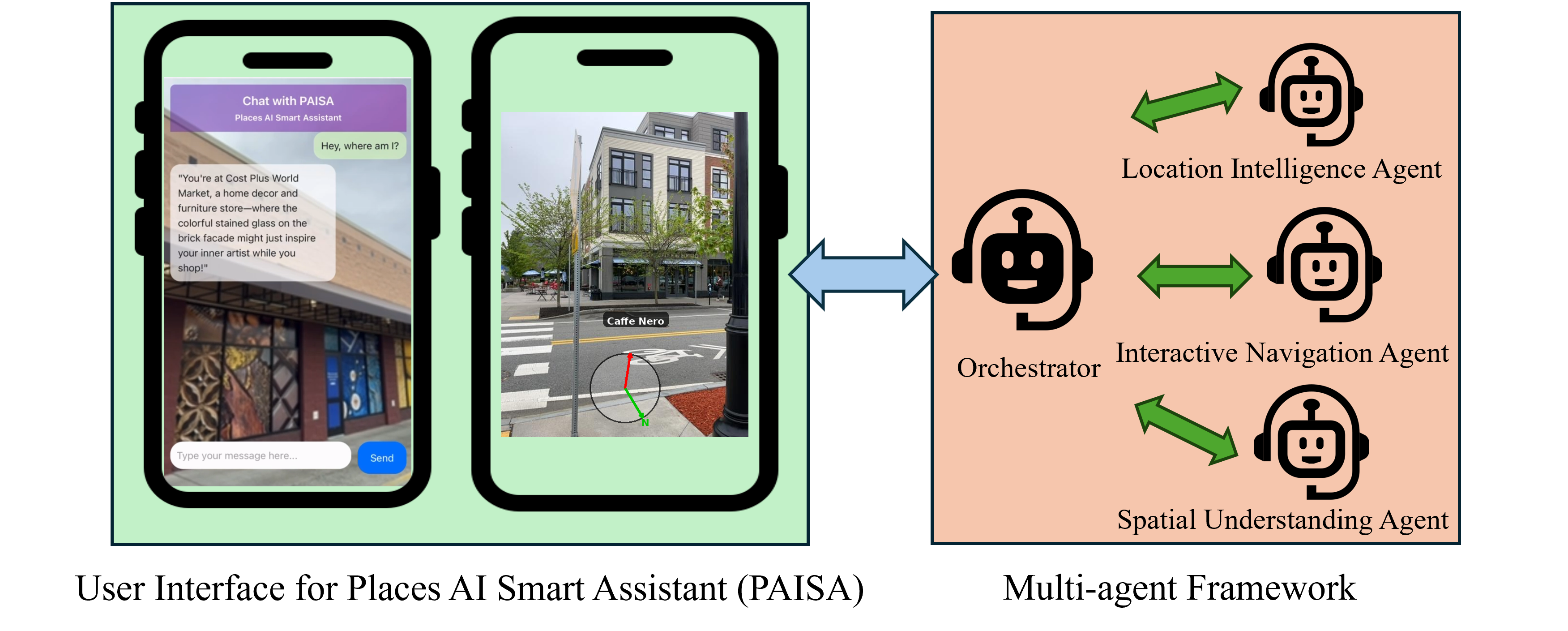}  % Adjust width as needed
    \caption{\textbf{The user interface (left) of the Places AI Smart Assistant and its underlying multi-agent framework (right).} PAISA offers two interface modes: a chatbot for answering user queries and an interactive navigation mode for destination guidance. The multi-agent framework consists of an orchestrator coordinating three specialized agents: the location intelligence agent, the interactive navigation agent, and the spatial understanding agent.}
    \label{fig:paisa_overall}
\end{figure*}
\subsubsection{Quadkey-based Visual Prompting}

Next, we convert the current map viewport into a quadkey-indexed~\cite{bingmaps_quadtile} grid and overlay this grid on a simplified map, as illustrated in Figure~\ref{fig:grid-layout}. Each quadkey tile is treated as a structured element in a \emph{visual prompt~\cite{yang2023setofmark}} to the multimodal LLM: the model is tasked with identifying tiles that contain salient map entities such as roads, parks, or water bodies. This quadkey-based discretization associates detected entities with precise spatial locations on the map, enabling spatial correlation analysis and reasoning over tile-level relationships. By segmenting the map into these visually prompted tiles, the model can decompose complex spatial layouts into manageable units, making downstream tasks---such as answering questions about specific regions or patterns across the viewport---more accessible and interpretable.

\subsubsection{Entity Search and Query Resolution}
When a user asks a question such as "What is the lake at the top right part of the map?", GPT-4o determines which part of the map they are referring to—such as the "top right" region—and retrieves relevant geographic entities using the Azure Maps API. The detected entities (e.g., Bonnet Lake, Abi’s Park) are appended to the user's query and reintroduced to GPT-4o for context-aware reasoning. The model then processes this enhanced prompt and provides a precise answer.
By integrating LLM-based reasoning with geospatial search capabilities, this system enables more intuitive interactions with maps, making it a valuable tool for travelers, researchers, and anyone exploring unfamiliar places.

\subsection{Places AI Smart Assistant (PAISA)}
Maps Plus provides a strong foundation for exploring geospatial search capabilities and reasoning.
However, it does not offer a natural way to interact with the real world or visually interpret the environment around the user. 
In practice, when someone is standing in front of an unfamiliar building or exploring new places in an unfamiliar city, relying solely on search-based tools is often insufficient. 
This gap motivates the need for a system that can seamlessly combine visual understanding with user geolocation, enabling richer, more intuitive, and context-aware interactions.
PAISA addresses this challenge by integrating multimodal signals—such as camera input and spatial context—to deliver a more immersive and informative real-world experience. 
The user interface of PAISA, illustrated in Figure~\ref{fig:paisa_overall}, is powered by a backend multi-agent system coordinated by an orchestrator agent. 
This system incorporates several specialized function agents, including a location intelligence agent, an interactive navigation agent, and a spatial understanding agent.
Each agent is powered by an LLM and equipped with a set of functional tools.

\subsubsection{Interactive Navigation Agent}
\begin{figure}[htbp]
    \centering
    \includegraphics[width=0.35\textwidth]{
    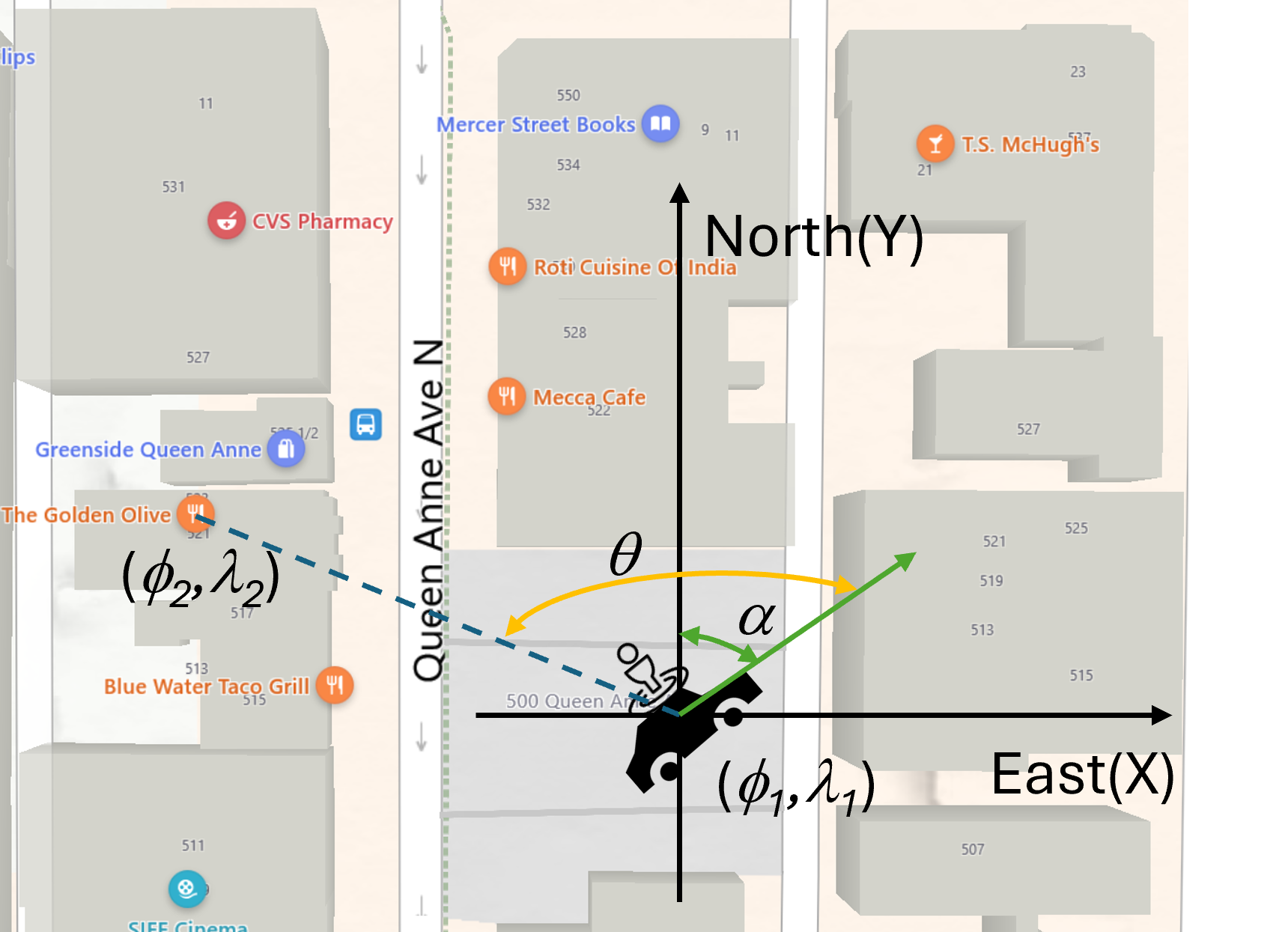}  % Adjust width as needed
    \caption{\textbf{Illustration of a person’s relative direction from the current position ($ \phi_1, \lambda_1$) to the destination($\phi_2, \lambda_2$).}}
    \label{fig:relative direction}
\end{figure}
\noindent The Interactive Navigation Agent (INA) is specifically designed to address the last-100-meter problem, helping users navigate the final stretch of their journey with precision. 
By leveraging the user's latitude, longitude, orientation, and destination coordinates, the agent guides users through the final segment, ensuring they reach their destination without confusion. The bearing to the destination is calculated using the formula:
\begin{align*}
\Delta \lambda &= \lambda_2 - \lambda_1 \\
\theta &= \arctan  \left( \sin(\Delta \lambda) \cdot \cos(\phi_2), \right. \\
    &\quad \left. \cos(\phi_1) \cdot \sin(\phi_2) - \sin(\phi_1) \cdot \cos(\phi_2) \cdot \cos(\Delta \lambda) \right)
\end{align*}
Where \( \phi_1, \phi_2 \) are latitudes and \( \lambda_1, \lambda_2 \) are longitudes of the user current location and destination location, and \( \Delta \lambda \) is the difference in longitude.
Next, the bearing \( \theta \) is adjusted for the user's orientation \( \alpha \) to find the relative direction (shown in Figure~\ref{fig:relative direction}):
\[
\text{Relative Direction} = \theta - \alpha
\]
Finally, ensure the direction is compass-friendly by adjusting for values outside the 0 to 360-degree range.
Additionally, INA includes a trigger feature that allows users to view the street view of their destination, offering a visual preview of the surroundings. 
This functionality enhances the user experience, providing clear and interactive navigation through the most challenging part of the journey, with real-time feedback and immersive, location-based guidance.
\subsubsection{Location Intelligence Agent}
\begin{figure}[htbp]
    \centering
    \includegraphics[width=\linewidth]{
    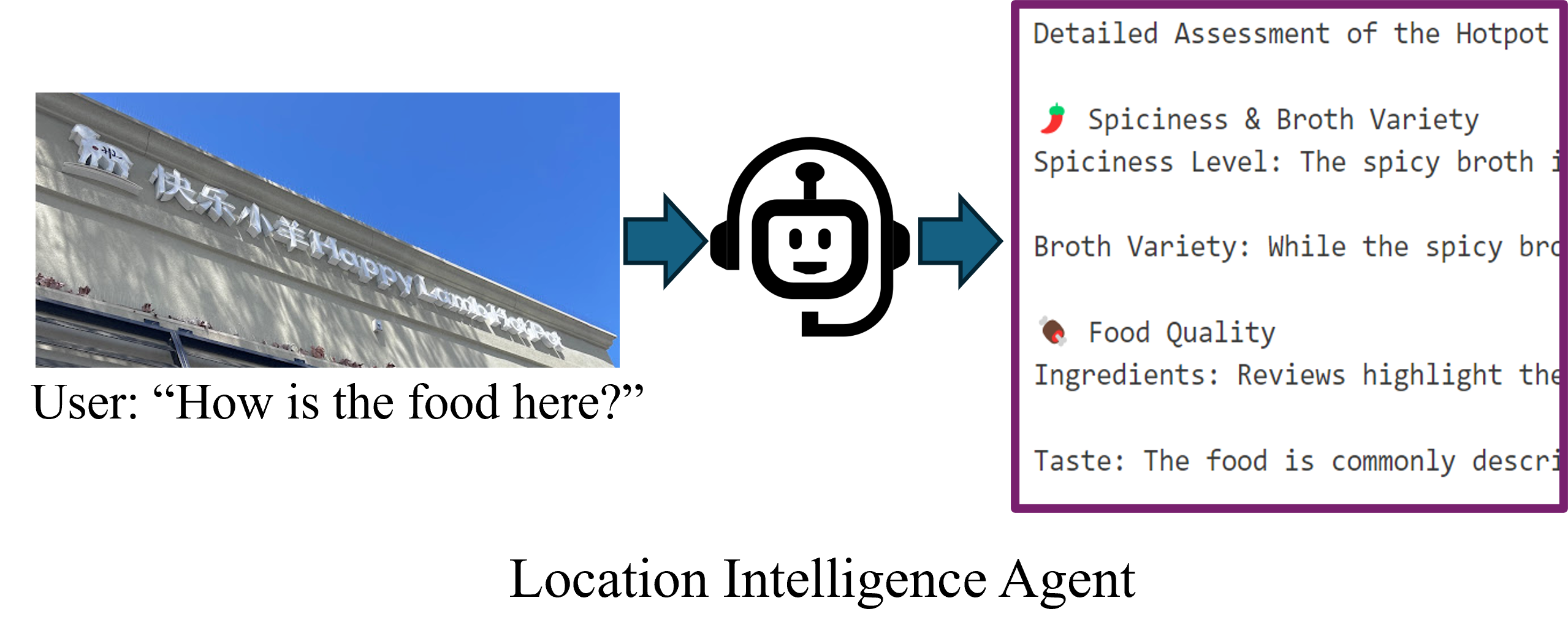}  % Adjust width as needed
    \caption{\textbf{Illustration of the location intelligence agent.} In this example, a user explores a new restaurant and inquires about its food; the agent identifies the place and integrates available information with user reviews to answer the query.}
    \label{fig:location_agent}
\end{figure}

The location intelligence agent enables users to explore unfamiliar places by identifying venues and retrieving relevant information to satisfy their curiosity. 
As illustrated in Figure~\ref{fig:location_agent}, this agent first determines what the place is and then leverages metadata and user reviews to enrich the understanding of that location. 
To ground a user-captured image to the correct venue, we encode the image using a Contrastive Language-Image Pre-training (CLIP)~\cite{radford2021learning} visual encoder, while each candidate place is represented with a CLIP text encoder applied to a structured descriptor that concatenates the place name, category, and latitude/longitude. 
From these representations, we construct a feature vector consisting of: (i) the cosine similarity between image and place embeddings, (ii) the distance between the user and the place, and (iii) a heading-consistency term defined as the absolute angular difference between the user’s device orientation and the bearing from the user to the place. 
To further enhance grounding accuracy, we augment this vector with local popularity indicators derived from Azure Maps search activity, providing data-quality priors. 
The resulting features are fed into an XGBoost ranking model, which assigns relevance scores and reorders the initial retrieval set. 
The top-ranked candidates are subsequently passed to a downstream LLM agent, supplying a compact, higher-recall context that improves the quality of the final answer.

\subsubsection{Spatial Understanding Agent}
Spatial reasoning is essential for connecting an ego-centric photo to nearby places during the final steps of navigation. Existing spatial VLMs (e.g., ASMv2~\cite{wang2024all}, SpatialVLM~\cite{chen2401spatialvlm}, SpatialRGPT~\cite{cheng2024spatialrgpt}) provide strong relational understanding but are often too slow or misaligned with the fine-grained urban cues required in our application. To obtain an efficient and task-adapted module that integrates cleanly into the IMAIA framework, we distill GPT\textendash4o into a Florence-2 model using a structured, multi-stage pipeline.

The agent takes a single street-level image and extracts (i) salient objects such as storefront signs, façades, or distinctive structural elements, and (ii) their spatial relationships, represented either as a scene graph or natural-language description. This representation supports two key system functions. First, when IMAIA retrieves cached street-level images of a destination, the agent generates relational descriptions (e.g., “the café entrance is just left of the red awning”), helping users visually confirm the location. Second, when users upload a photo, the agent converts it into a structured spatial record that downstream components use for grounding, matching, and contextual disambiguation. This pairing of destination-side and user-side spatial grounding improves interpretability and reliability in last-meter navigation.

Our distillation pipeline consists of three stages. In stage (i), GPT\textendash4o-mini is prompted to extract candidate key entities for 40k street-view images, and the most frequent elements are retained as salient anchors. In stage (ii), YOLO-World~\cite{cheng2024yolo} and Depth Anything V2~\cite{depth_anything_v2} provide 2D localization and depth cues, and GPT\textendash4o generates pairwise spatial relations using Set-of-Mark-style prompting. In stage (iii), we build a supervised fine-tuning set by pairing each annotated image with diverse relational queries reflecting real urban navigation needs. We then fine-tune a Florence-2 model in its dense captioning format, yielding a compact and responsive spatial reasoning module suitable for real-time deployment within IMAIA.

\subsubsection{Handle complex query with multi-agent reasoning}
\begin{figure}[htbp]
    \centering
    \includegraphics[width=\linewidth]{
    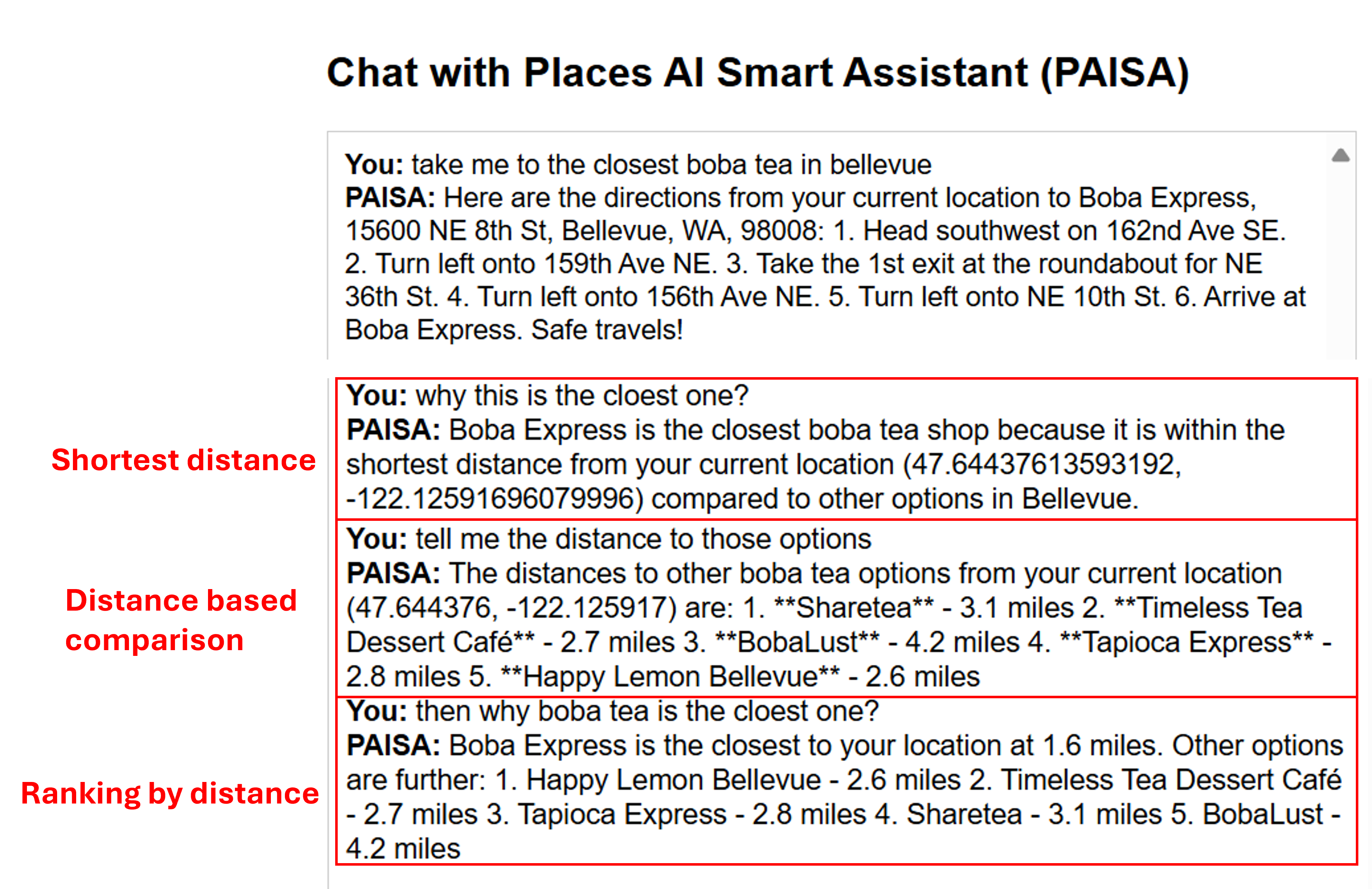}  % Adjust width as needed
    \caption{\textbf{Example of reasoning with the PAISA.} The system answers user queries by reasoning over geospatial information: (top) providing directions to the nearest boba tea shop, (middle) justifying the choice based on shortest distance and comparative options, and (bottom) ranking alternatives by distance.}
    \label{fig:reasoning}
\end{figure}

\begin{figure}[htbp]
    \centering
    \includegraphics[width=\linewidth]{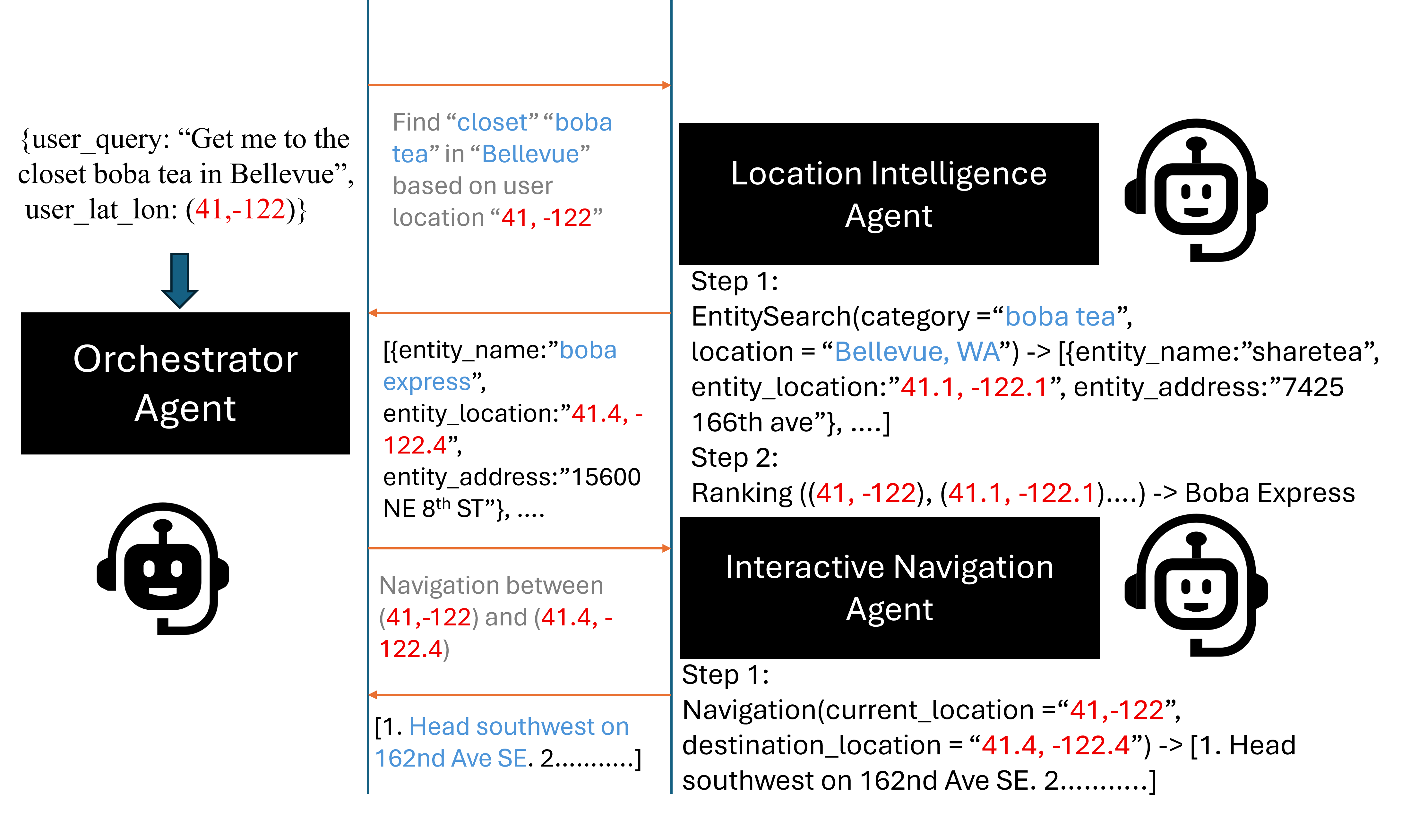}  % Adjust width as needed
    \caption{\textbf{An example of backend multi-agent workflow of PAISA.} The orchestrator agent parses the user’s query and delegates tasks to specialized agents: the location intelligence agent identifies the relevant place (e.g., Boba Express in Bellevue), while the interactive navigation agent generates turn-by-turn directions to the destination.}
    \label{fig:multi-agent}
\end{figure}

Modern map applications often fall short when handling complex, user-centric queries. 
Consider a scenario where a user urgently seeks the nearest boba tea shop. 
Executing this seemingly simple request typically involves multiple steps: initiating a search for boba tea within a specified area, manually reviewing and ranking results based on proximity, and finally selecting a navigation option to begin the route. 
This fragmented interaction model demands several discrete actions from the user, rather than supporting a seamless, single-command experience—highlighting a common usability limitation in current systems.
To address this challenge, we introduce a multi-agent framework capable of interpreting and executing such queries end-to-end. 
As illustrated in Figure~\ref{fig:reasoning}, our proposed system, PAISA, directly outputs navigation instructions to the closest boba tea shop based on the user’s location. 
Naturally, this raises the question of whether such output is a hallucination from a large language model (LLM). 
To validate the reasoning, we examine PAISA’s internal decision process: the system performs entity search within the target area, computes distances between the user and candidate locations, and ranks them accordingly. 
In this case, “Boba Express” emerges as the top result due to its proximity—just 1.6 miles from the user—demonstrating the system’s grounded and interpretable reasoning.
The backend of PAISA is organized as a multi-agent system (Figure~\ref{fig:multi-agent}). 
The orchestrator agent first analyzes the query and decomposes it into simpler sub-queries. 
These are passed to the location intelligence agent, which retrieves candidate entities and their attributes. 
The orchestrator then forwards the enriched information to the interactive navigation agent, which generates the optimal route. 
Finally, the navigation plan is returned to the orchestrator and delivered to the user.
\section{Results}
\label{sec:results}
\subsection{Place detection accuracy with Maps Plus}
In this section, we evaluate and compare the quality of geospatial entity search across several methods. Specifically, we benchmark the Maps Plus approach against three widely used baselines (Figure~\ref{fig:maps_plus_comp}) in Table~\ref{tab:maps_plus_comp}: (1) \textit{Single Model}: The model receives only the user query and the current map view image, relying solely on its pretrained knowledge. (2) \textit{Model + Location}: The input includes the query, the map view image, and the geographic coordinates (latitude and longitude) of the map view. (3) \textit{Model with Verbose Location}: Similar to the previous setup, but the location input is replaced with verbose descriptors such as city names and landmarks.
In this study, we construct a dataset by selecting 10 cities across the United States and sampling points of interest (POIs) within a 20-kilometer radius of each city center. For each POI, we employ GPT-4o to generate synthetic queries based on information such as the POI’s attributes, geographic coordinates, and related contextual data. This process yields a total of 4,300 queries, for example, \textit{What is the lake at the top left part of the map}.
\begin{table}[htbp]
    \centering
    \caption{\textbf{POI detection accuracy across different methods illustrated in Figure~\ref{fig:maps_plus_comp}.} The proposed method significantly improves performance using the same LLM backbone.}
    \label{tab:maps_plus_comp}
    \begin{tabular}{lc}
        \toprule
        \textbf{Method} & \textbf{Accuracy} \\
        \midrule
        Single Model              & 39.30\% \\
        Model + Location          & 41.46\% \\
        Model + Verbose Location  & 42.74\% \\
        Maps Plus           & 89.83\% \\
        \bottomrule
    \end{tabular}
\end{table}
\begin{figure}[htbp!]
    \centering
    \includegraphics[width=1\linewidth]{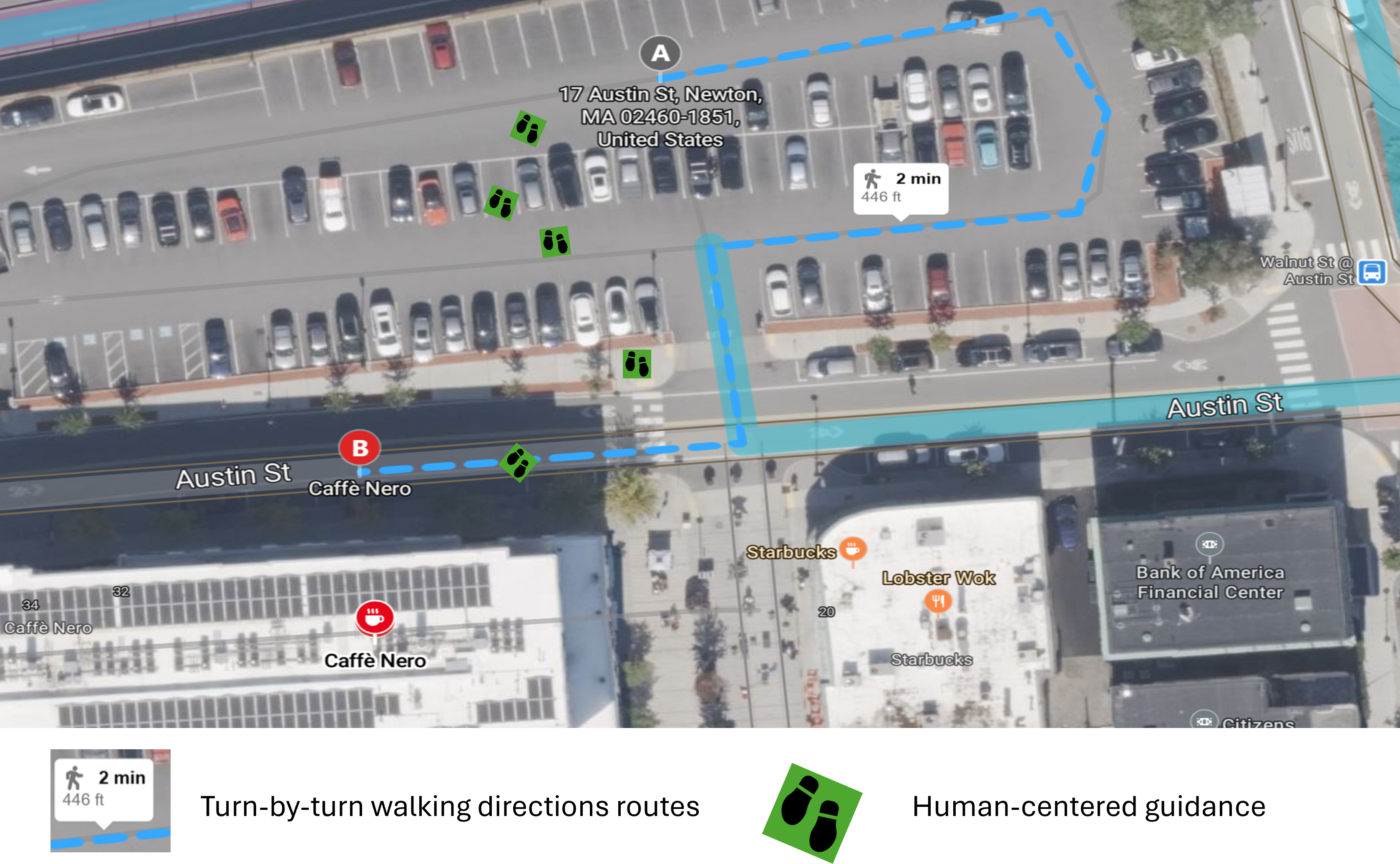}  % Adjust width as needed
    \caption{\textbf{Comparison of human-centered guidance vs. conventional turn-by-turn walking directions to the destination (Caffe Nero).} Turn-by-turn navigation follows a fixed path derived from map topology, which can introduce unnecessary detours, whereas the human-centered approach interactively points the user toward the destination using real-time relative direction, reducing extra walking.}
    \label{fig:daa_exp_1}
\end{figure}

\begin{figure}[htbp!]
    \centering
    \includegraphics[width=1\linewidth]{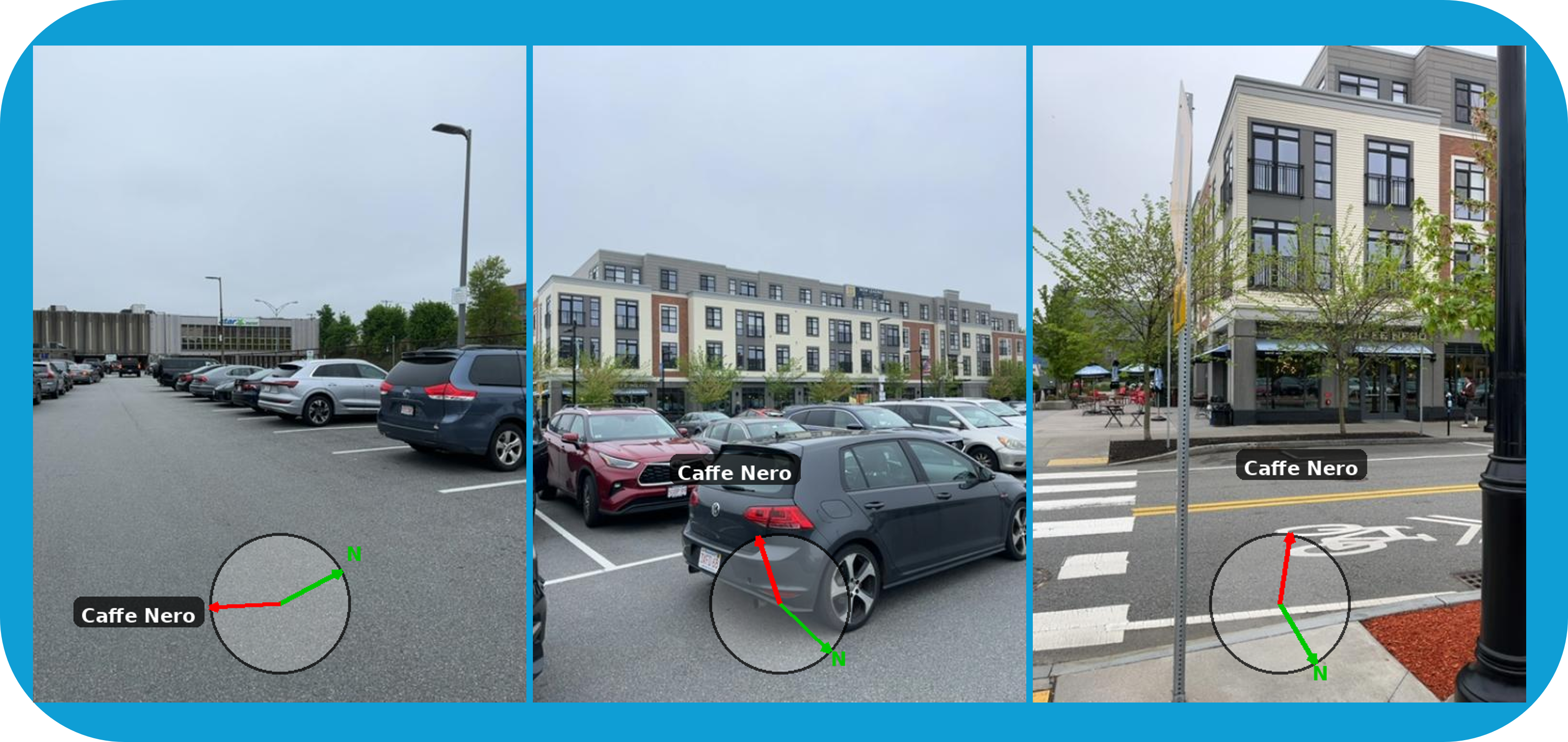}  % Adjust width as needed
    \caption{\textbf{Interactive navigation UI of PAISA guiding a pedestrian to Caffe Nero.} Sequential views from the parking lot to the storefront. The circular overlay shows the real-time bearing to the destination (red arrow) relative to north (green), enabling direct, flexible guidance.}
    \label{fig:daa_exp_2}
\end{figure}

Table~\ref{tab:maps_plus_comp} presents the accuracy achieved by various methods. The results indicate that our proposed approach attains significantly higher accuracy compared to the baselines, without requiring any fine-tuning of the LLM. The superior performance of the proposed method can be attributed to its efficient integration of grounding data, specifically entities retrieved from geospatial services, into the LLM's processing. For instance, consider the query \textit{"What is the coffee shop below the cinema?"}. The proposed method not only furnishes a pertinent set of local entities but also deconstructs the query into a series of sub-problems. This multi-step decomposition facilitates improved LLM reasoning and consequently, enhanced performance.
% \begin{table*}[htbp!]
% \centering
% \begin{tabular}{@{}ccc@{}}
% \toprule
% Distance (meter)        & Average Match Accuracy & Runtime Performance (seconds) \\ \midrule
% 100          & 84.53\%                & 0.54s               \\
% 150          & 82.47\%                & 1.37s               \\
% 200          & 81.14\%                & 2.71s               \\
% 300          & 71.25\%                & 5.01s               \\
% 500          & 64.31\%                & 15.19s              \\
% 1000         & 53.42\%                & 71.22s           
% % \midrule
% % No filtering & $\sim$0\%              & $\infty$              
% \\ \bottomrule
% \end{tabular}
% \caption{Results of comparing the distance parameter in embedding-based entity search. }
% \label{tab:filter_result}
% \end{table*}

\begin{table*}[htbp!]
    \centering
    \caption{Performance of venue candidate ranking methods in terms of Precision and Recall at Top-1 and Top-3.}
    \label{tab:ranker_perf}
    \begin{tabular}{lcccc}
        \toprule
        \textbf{Ranker} & \textbf{Precision@Top-1} & \textbf{Recall@Top-1} & \textbf{Precision@Top-3} & \textbf{Recall@Top-3} \\
        \midrule
        XGBoost Ranker     & 80.4\% & 72.5\% & 36.2\% & 92.8\% \\
        Distance-sorting   & 76.1\% & 69.2\% & 30.4\% & 77.5\% \\
        Similarity-sorting & 65.2\% & 58.3\% & 25.4\% & 68.1\% \\
        \bottomrule
    \end{tabular}
\end{table*}

\subsection{Comparison between human-centered path and turn-by-turn walking directions path}
We compare human-centered guidance with conventional turn-by-turn (TBT) navigation (Figure~\ref{fig:daa_exp_1}). TBT routes rely on map topology and predefined road graphs, which may introduce unnecessary detours. In contrast, pedestrians often choose more direct paths, such as crossing open spaces or taking informal shortcuts, revealing a gap between algorithmic routing and human spatial reasoning. Human-centered guidance addresses this limitation by directing users toward the destination using their first-person camera view and orientation. As illustrated in Figure~\ref{fig:daa_exp_2}, our system integrates camera input, geolocation, and device orientation to compute destination bearings and overlay AR guidance cues. We evaluate this approach on 10 last-100-meter navigation scenarios. Four scenarios require a turn where the destination is not directly visible, and six involve destinations that are visible but may be temporarily occluded. Walking time is measured relative to standard TBT navigation. In turn-required scenarios, human-centered guidance reduces average walking time from 3.28 (min) to 2.08 (min) (63.5\% of TBT). In directly visible scenarios, the reduction is larger, from 3.36 (min) to 1.07 (min) (32.1\% of TBT). These results indicate that bearing-aware, scene-informed guidance can significantly reduce detours and improve navigation efficiency. The interface and interaction design were further refined through informal feedback from approximately 15 users, improving clarity and usability and resulting in the final interface presented in the paper.% We compare the human-centered path to the turn-by-turn walking directions path, as illustrated in Figure~\ref{fig:daa_exp_1}. 

\subsection{Evaluation on the embedding-based entity search}
% We construct our review database using a combination of proprietary review data and publicly available Yelp review data \cite{alam2021yelp}.  
% To evaluate the performance of our system, we use a collection of test images sourced both locally and from online platforms. 
% Our evaluation spans three different cities, each with multiple locations, where we compute the average matching accuracy between the test images and the database entries. 
% Given a user’s geolocation, we define a filtering distance to constrain the candidate review entities considered during the matching process. 
% Specifically, when the distance is set to 100 meters, the system attempts to match the input image with review images associated with locations within that spatial boundary.
% As shown in Table~\ref{tab:filter_result}, we observe that the average match accuracy decreases as the filtering distance increases. 
% Within a 200-meter distance, the system maintains an accuracy above 80\%; however, accuracy drops significantly to approximately 50\% when the distance expands to 1,000 meters. 
% This trend suggests that spatial locality plays a critical role in the reliability of visual matching. 
% Additionally, the system’s runtime increases with larger distance due to the growing number of candidate entities. 
% These results underscore the importance of leveraging geospatial filtering constraints to balance accuracy and efficiency in location-aware image-based entity retrieval systems.
The XGBoost ranker was trained on a dataset of 500 image queries, each paired with manually annotated ground‑truth venues, and evaluated on a held‑out set of 50 queries. For benchmarking, we considered two baselines: (i) a distance‑based method that orders candidate places solely by geodesic proximity to the user, and (ii) a similarity‑based method that ranks candidates exclusively according to the cosine similarity between image and place embeddings. Table~\ref{tab:ranker_perf} summarizes the comparative performance of the proposed XGBoost ranker and two baseline methods, evaluated using Precision and Recall at Top‑1 and Top‑3. The Top-k Precision and Recall are defined as:
\begin{align*}
\text{Precision@}k &= \frac{\left| \{ \text{relevant items in top-}k \} \right|}{k}, \\[6pt]
\text{Recall@}k &= \frac{\left| \{ \text{relevant items in top-}k \} \right|}{\left| \{ \text{all relevant items} \} \right|}.
\end{align*}
Top‑1 metrics ($k = 1$) measure whether the highest‑ranked candidate is correct, providing a strict indicator of ranking accuracy at the very top. In contrast, Top‑3 metrics ($k = 3$) evaluate the proportion and coverage of relevant items within the first three positions, reflecting the system’s ability to surface multiple correct candidates early in the ranking. As shown in Table~\ref{tab:ranker_perf}, the XGBoost ranker consistently outperforms both distance‑based and similarity‑based baselines across all metrics, with the largest gains observed in Top‑3 recall, indicating improved breadth of relevant retrieval without sacrificing top‑rank precision.

\subsection{Evaluation on Spatial Understanding }
\begin{figure}[htbp]
    \centering
    \includegraphics[width=\linewidth]{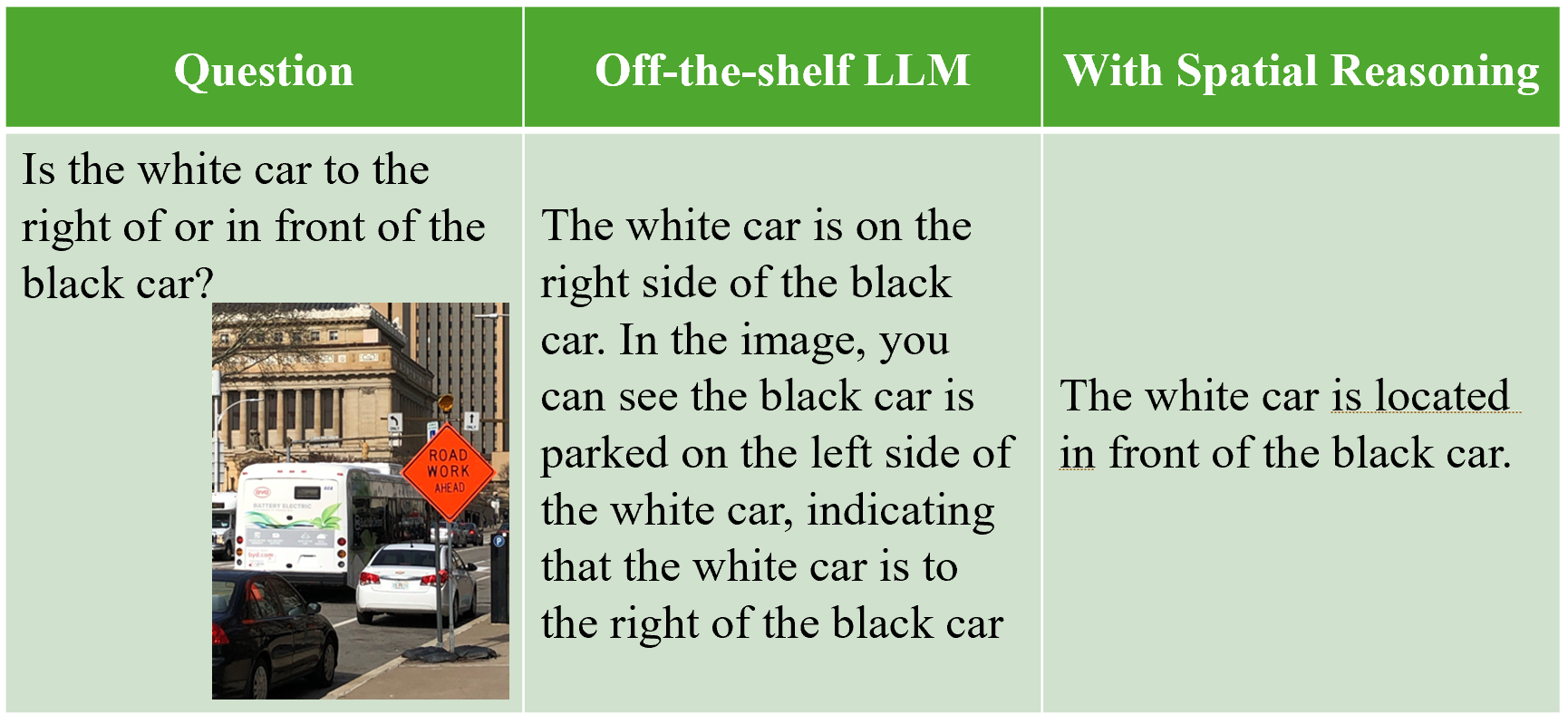}  % Adjust width as needed
    \caption{\textbf{Example result of our proposed method compared with Off-the-shelf LLM}.}
    \label{fig:spatial_understanding}
\end{figure}

We evaluate the proposed spatial reasoning module on a test set of 400 street-view images, comparing against both general-purpose multimodal LLMs and specialized spatial-scene graph models. Accuracy is measured with an LLM-as-judge protocol using the OpenAI o1 model, while efficiency and recall are assessed with task-specific metrics. 
\textbf{Comparison with multimodal LLMs.}  
As shown in Figure~\ref{fig:spatial_understanding}, our distilled model achieves an accuracy of 84\%, indicating that the majority of generated spatial descriptions are judged as correct by the o1 evaluator. In contrast, Florence-VL 8B~\cite{chen2025florence}, a general multimodal LLM built upon Florence-2 with nearly ten times more parameters, attains only 27\% accuracy under the same setting. This result highlights the effectiveness of task-aligned distillation for spatial reasoning compared to parameter scaling. 
\textbf{Comparison with scene-graph models.}  
Against specialized spatial reasoning systems such as ASM v2~\cite{wang2024all}, which can generate structured scene graphs but lack natural language interaction capabilities, our model demonstrates higher recall of salient items. On average, our model identifies approximately 7 objects per scene, compared to 4 objects extracted by ASM v2. This improvement suggests that combining structured spatial grounding with natural language reasoning enables richer scene interpretation. 
\textbf{Comparison with agent-based solutions.}  
We further benchmark against an agentic pipeline that replicates Stage (i) and (ii) with explicit calls to external models. On a single NVIDIA V100 32GB GPU, the agent-based approach requires 12.4s per image, while our end-to-end distilled model reduces inference time to 1.7s per query, achieving a 7.3$\times$ speedup. This efficiency gain is critical for real-time deployment in navigation scenarios. 
Overall, these results demonstrate that our distilled Florence-2 model achieves strong accuracy, improved recall, and significant efficiency gains over both large-scale multimodal LLMs and agent-based baselines, making it well-suited for spatial reasoning in real-world.

\section{Conclusion}
In this work, we introduce IMAIA, an interactive Maps AI Assistant that spans desktop-style map exploration and on-the-ground, last–100–meter navigation in a unified, deployment-oriented framework. IMAIA is built around two complementary components---Maps Plus for map-centric spatial understanding and PAISA for ego-centric scene understanding and geospatial place grounding---coordinated by a lightweight multi-agent orchestration layer. Maps Plus treats the map viewport as a quadkey-indexed visual prompt, enabling view-conditioned, map-centric queries and boosting place detection accuracy from under 43\% with conventional approaches to nearly 90\%. PAISA operates in an AR-style setting, fusing camera input with geospatial signals to provide human-centered, bearing-aware guidance that better matches natural pedestrian behavior than rigid turn-by-turn instructions, reducing unnecessary detours. Beyond individual modules, our results show that IMAIA’s model-agnostic, modular design supports efficient and practical geospatial reasoning for user-facing applications. A distilled spatial reasoning model within the framework attains 84\% accuracy and a 7.3$\times$ inference speedup over agent-based pipelines (1.7s vs.\ 12.4s), demonstrating the effectiveness of lightweight, task-aligned distillation. Together, these findings indicate that focusing on system design--quadkey-based visual prompting for maps, AR-style camera grounding, and a replaceable VLM/vision backend enables an interactive and responsive geospatial experience and production-ready quality for trip planning and local discovery.
{
    \small
    \bibliographystyle{plainnat}
    \bibliography{main}
}

% WARNING: do not forget to delete the supplementary pages from your submission 
% \input{sec/X_suppl}

\end{document}